\documentclass[]{fairmeta}
\usepackage{amsthm,amsmath,amssymb}
\usepackage{graphicx}
\usepackage{natbib}
\usepackage{subcaption}
\newtheorem{theorem}{Theorem}[]

\newtheorem{remark1}[theorem]{Remark}

\title{TeleWorld: Towards Dynamic Multimodal Synthesis with a 4D World Model}
\author{TeleWorld Team}
\metadata[Corresponding Author]{Xuelong Li(\email{xuelong$\_$li@ieee.org})}

\begin{document}

\abstract{
World models aim to endow AI systems with the ability to represent, generate, and interact with dynamic environments in a coherent and temporally consistent manner. While recent video generation models have demonstrated impressive visual quality, they remain limited in real-time interaction, long-horizon consistency, and persistent memory of dynamic scenes, hindering their evolution into practical world models. In this report, we present TeleWorld, a real-time multimodal 4D world modeling framework that unifies video generation, dynamic scene reconstruction, and long-term world memory within a closed-loop system. TeleWorld introduces a novel generation-reconstruction-guidance paradigm, where generated video streams are continuously reconstructed into a dynamic 4D spatio-temporal representation, which in turn guides subsequent generation to maintain spatial, temporal, and physical consistency. To support long-horizon generation with low latency, we employ an autoregressive diffusion-based video model enhanced with Macro-from-Micro Planning (MMPL)--a hierarchical planning method that reduces error accumulation from frame-level to segment-level-alongside efficient Distribution Matching Distillation (DMD), enabling real-time synthesis under practical computational budgets. Our approach achieves seamless integration of dynamic object modeling and static scene representation within a unified 4D framework, advancing world models toward practical, interactive, and computationally accessible systems. Extensive experiments demonstrate that TeleWorld achieves strong performance in both static and dynamic world understanding, long-term consistency, and real-time generation efficiency, positioning it as a practical step toward interactive, memory-enabled world models for multimodal generation and embodied intelligence.}

\maketitle

\vspace{-0.1em}

\section{Introduction}

The pursuit of artificial intelligence systems capable of understanding, simulating, and interacting with the physical world has driven significant progress in world modeling research~\cite{ding2025understandingworldpredictingfuture, zhu2025soraworldsimulatorcomprehensive}. Recent advances have demonstrated the promise of world models, highlighting that explicit reconstruction of the world and real-time generation are complementary and mutually reinforcing capabilities. At their core, world models aim to endow AI systems with human-like perception and interaction abilities—enabling machines to not only observe and represent dynamic environments but also to predict, generate, and meaningfully engage with them in real time.~\cite{guo2025ctrlworldcontrollablegenerativeworld, xiao2025worldenvleveragingworldmodel, guo2025mineworldrealtimeopensourceinteractive, Zuo_2025_CVPR, won2025dualstreamdiffusionworldmodelaugmented, wang2025videoversefart2vgenerator, hu2025text2worldbenchmarkinglargelanguage, jin2025occtens3doccupancyworld, chen2025geodrive3dgeometryinformeddriving}

Let's start by discussing what a world model is. The definition of a world model varies across research communities~\cite{ding2025understandingworldpredictingfuture, zhu2025soraworldsimulatorcomprehensive}, reflecting the multifaceted nature of this emerging field. Different researchers have varying interpretations, but broadly speaking, world models encompass several interconnected research directions, including but not limited to video generation~\cite{wang2025videoversefart2vgenerator}, 3D reconstruction~\cite{Zuo_2025_CVPR}, embodied AI~\cite{guo2025ctrlworldcontrollablegenerativeworld}, and autonomous driving~\cite{chen2025geodrive3dgeometryinformeddriving,jin2025occtens3doccupancyworld}. In a general sense, any model that can naturally represent the world and interact with it may be considered a world model. However, the video generation direction has become a more popular research area within the field of world models, thanks to its higher quality output, stronger downstream multi-task capabilities (i.e., results from video generation can also be applied in areas such as embodied AI and autonomous driving), and its greater accessibility and interactivity for users.

However, video generation models themselves have several fundamental shortcomings that hinder their evolution into more practical world models~\cite{Causvid, SelfForcing, xiang2025MMPL}. First, due to the structural limitations of multi-step denoising pipelines in video diffusion models, video generation is heavily restricted in meeting the real-time generation and interaction requirements of a world model. Second, long-term video generation still faces significant challenges with temporal consistency over extended durations. Extended world exploration and interaction often suffer from error accumulation and quality degradation. Third, a world model needs to retain a certain memory of the generated world, which is inherently four-dimensional—spanning the three dimensions of space and the dynamic dimension of time, just as human perception of the world. Existing world models and video generation approaches often only capture memory from past video sequences or three-dimensional representations, while achieving four-dimensional memory remains a significant difficulty in the video generation path toward world models. Finally, high-quality video generation models are typically computationally expensive, making it difficult to train and deploy them in a fast, efficient, and sustainable manner with real-time capability. The hardware demands for world models following the video generation approach remain prohibitively high for many researchers.

Here we summarize the key issues as: (1) \textbf{Modeling Dynamic 4D Scenes}: Current world models, which primarily possess 3D modeling, struggle to effectively model and memorize dynamic environments with full spatio-temporal coherence. (2) \textbf{Ensuring Long-term Consistency}: Maintaining both high fidelity and temporal consistency over extended generation periods remains difficult, often leading to issues like color shift and quality degradation. (3) \textbf{Balancing Real-time Efficiency with Quality}: Achieving real-time generation and efficient training is a primary challenge, as it requires reconciling high model quality with manageable computational cost.

In this report we propose TeleWorld, a practical real-time 4D world model that addresses these fundamental challenges through a unified framework integrating generation, reconstruction, and guidance. Firstly, we propose a "Generation-Reconstruction-Guidance" closed-loop paradigm that records the dynamic scene during video generation using a 4D spatio-temporal field. This reconstruction process runs synchronously with generation, continuously updating the world representation as new content is synthesized. The rendering results of this 4D field are then used as guidance to steer subsequent generation, ensuring spatial consistency, temporal coherence, and physical plausibility. This reconstruction-based approach achieves long-term dynamic memory through persistent, coherent understanding of the generated world. During the generation stage, we employ an autoregressive diffusion video generation model equipped with planning capabilities. Drawing insights from recent advances in planning-based generation, our Macro-from-Micro Planning (MMPL) framework~\cite{xiang2025MMPL} operates hierarchically: micro-planning predicts key anchor frames within short video segments to establish local temporal coherence, while macro-planning chains these segments autoregressively to achieve global consistency across long horizons. This approach reduces error accumulation from the frame level to the segment level, enabling stable, high-quality generation over extended durations. Our video generation architecture enables faster video synthesis while allowing better integration of information from the 4D scene during the planning process. 

To further accelerate video synthesis, we adopt Distribution Matching Distillation (DMD)~\cite{yin2024onestepdiffusiondistributionmatching} on top of TeleWorld. While DMD is critical for real-time video generation, applying it to an auto-regressive model with more than 10B parameters is highly non-trivial, as it simultaneously introduces a large KV cache and requires working with three 10B-plus models—the generator, teacher, and critic. Even with Fully Sharded Data Parallelism (FSDP~\cite{zhao2023pytorchfsdpexperiencesscaling}), this combined memory footprint exceeds the capacity of 64 NVIDIA H100 GPUs~(\cite{krea_realtime_14b}).
To address this challenge, we propose a novel training system for large-scale Distribution Matching Distillation. Specifically, we assign the generator, teacher, and critic to disjoint sets of GPUs and orchestrate their execution using Ray~(\cite{PhilippRay}). In addition, we employ context parallelism to shard the generator’s KV cache across devices, substantially reducing per-GPU memory consumption. 
Furthermore, we carefully design a pipeline execution schedule that minimizes GPU idle time (i.e., pipeline bubbles) and improves overall training efficiency. With these optimization techniques, we successfully train DMD for Teleworld-18B using only 32 H100 GPUs. Together, these system-level optimizations enable real-time video generation with modest training overhead under practical computational budgets.

Through these innovations, TeleWorld achieves seamless integration of dynamic object modeling and static scene representation within a coherent 4D framework, advancing world models toward practical, interactive, and computationally accessible systems suitable for multimodal generation and embodied intelligence applications.

The contributions of our method can be summarized as follows:
\begin{itemize}
    \item We propose a real-time ``generation–reconstruction–guidance'' closed-loop framework that reconstructs long-term memory from the world model into dynamic point clouds at real-time speed while maintaining rapid world updates and temporal consistency.
    \item We introduce a dynamic four-dimensional world model that not only provides memory and generation capabilities in three-dimensional space but also enables the memorization and generation of moving objects within the scene, achieving true spatio-temporal coherence.
    \item We propose a novel training system that unlocks distillation training of large-scale autoregressive diffusion models, allowing efficient training on accessible hardware configurations while enabling real-time generation capabilities without compromising model quality.
    \item TeleWorld represents a comprehensive approach to world modeling that bridges video generation, 3D reconstruction, and persistent memory within a single unified system, positioning it as a practical foundation for interactive AI systems and embodied intelligence applications.
\end{itemize}

\section{Related Works}

\subsection{World Models}
%
The question of what constitutes a world model is a topic of frequent discussion among researchers today. The mainstream discussion centers on the idea that a world model is, in essence, an environment that can be navigated and interacted with. Consequently, much of the research has focused on how to construct such an environment. With the rise of generative models in recent years, world models have gradually branched into two main categories: 3D-based world models and video-based world models. The former first constructs a three-dimensional world and then renders it for the user, while the latter builds the world through video generation.

\textbf{3D-based World Models} A notable example in this field is Wonderworld~\cite{Yu_2025_Wonderworld}, which exhibits the ability to produce interactive 3D environments from just one 2D image. This highlights the possibility of building navigable virtual worlds with very limited initial data. The methodology prioritizes maintaining spatial coherence, accurate geometric interpretation, and low-latency feedback for user movement and actions.

Progress in this area has since broadened these functionalities. For instance, Matrix-3D~\cite{yang2025matrix3domnidirectionalexplorable3d} accomplishes extensive, all-directional 3D world creation that users can explore, utilizing panoramic 3D reconstruction techniques. Meanwhile, HunyuanWorld 1.0~\cite{hunyuanworldteam2025hunyuanworld10generatingimmersive} delivers fully immersive $360^{\circ}$ environments by employing semantically structured 3D mesh models, ensuring smooth integration with standard computer graphics workflows. In parallel, World Labs has entered the commercial space with its first product, Marble. This multimodal world model can create high-fidelity, persistent 3D worlds from a single image, video clip, or text prompt. The company differentiates itself by focusing on generating persistent, downloadable 3D environments.

\textbf{Video-based world models} For instance, Cosmos~\cite{nvidia2025cosmosworldfoundationmodel} has achieved breakthrough performance in realistic simulations for robotics and autonomous systems. Meanwhile, Genie 3~\cite{genie3} has introduced real-time interaction capabilities, allowing users to generate and navigate controllable 3D worlds with high consistency. In contrast, models such as Hunyuan-Voyager~\cite{Voyager}, which outputs 3D point clouds via RGB-D video, Hunyuan-GameCraft2~\cite{tang2025hunyuangamecraft2instructionfollowinginteractivegame} designed for game videos with hybrid historical conditioning, and Adobe‘s RELIC~\cite{hong2025relicinteractivevideoworld}, which employs a compact KV cache for long-term memory prioritize explicit 3D consistency and spatial reconstruction. The video-based approach offers distinct advantages for dynamic, user-centric applications: it delivers higher perceptual quality in motion and temporal coherence, supports more intuitive and responsive interaction due to its frame-by-generation nature, and enables rapid "cold-start" scene expansion—effectively allowing seamless "dream-outward" extension from minimal initial inputs. 

However, current video-based world models are primarily limited to handling static 3D environments and often struggle to effectively model dynamic objects within these worlds.


\subsection{Real-time Video Generation}
Recent advances in long‑video generation have largely been driven by autoregressive diffusion models~\cite{MAGI}. Techniques such as Causvid~\cite{Causvid} and Self‑Forcing~\cite{SelfForcing} have been introduced to improve training stability and temporal coherence by conditioning each new frame on previously generated content. While these methods can produce extended sequences, they remain susceptible to error propagation over long horizons, where small inconsistencies in early frames gradually amplify and degrade visual quality. Moreover, maintaining long‑range temporal consistency remains a fundamental challenge—models often “forget” earlier scene geometry or object identities, leading to incoherent narratives or visual artifacts in longer generations.

In parallel, real‑time video generation has aimed to deliver low‑latency, interactive synthesis. Yet scaling such systems to high‑quality, high‑resolution output—especially with large‑parameter models such as 10B-plus architectures—presents significant difficulties. The real‑time distillation of such models is particularly demanding, as it requires compressing both spatial and temporal knowledge without sacrificing fidelity, while also managing severe computational and memory constraints during deployment. Although \cite{krea_realtime_14b} overcomes this challenge on a 14B model with dynamic KV cache management, the fundamental problem is not solved without sharding the KV cache. 

These challenges—error accumulation, long‑term memory decay, and the difficulty of distilling high‑quality large models for real‑time use—motivate the design of TeleWorld. Instead of relying solely on implicit neural representations or recurrent latent states, TeleWorld introduces an explicit 4D spatiotemporal field that continuously records and reconstructs the evolving world. This explicit representation preserves geometric and appearance information across time, effectively mitigating common failure modes such as forgetting and inconsistency, while enabling efficient, high-fidelity long-video generation and real-time inference. Moreover, thanks to the use of context parallelism to shard the generator’s KV cache across devices, TeleWorld-18B can be trained for long-video distillation using only 32 H100 GPUs.


\section{Methods}


\subsection{``Generation-Reconstruction-Guidance'' Loop}

\begin{figure*}[t]
    \centering
    \includegraphics[width=0.8\textwidth]{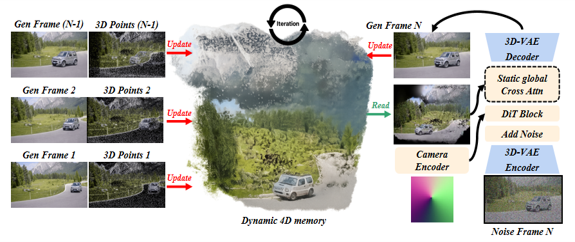}
    \caption{Structure of TeleWorld. We propose a dynamic "Generation-Reconstruction-Guidance" closed-loop framework for 4D spatio-temporal modeling. The model first generates an initial set of videos based on the user’s pre-defined instructions. It then enters a loop where, in each iteration, it processes the user’s real-time input instructions, reconstructs the video output from the previous round, and renders it according to the input camera poses. The rendered results serve as guidance to direct the current round of video generation and motion synthesis, and this process repeats iteratively.}
    \label{fig:structure}
\end{figure*}

We introduce a dynamic “Generation-Reconstruction-Guidance” closed-loop framework for unified 4D spatial–temporal modeling. This framework constructs a real-time, native 4D world representation that updates continuously with each newly generated video segments, ensuring perfect synchronization with the evolving visual content. A core innovation is the seamless alignment between dynamic object modeling and static scene modeling, enabling their unified integration within a coherent spatial structure. In this loop, reconstruction refers to the process of recovering a consistent 4D scene representation from generated frames, while guidance denotes the use of both the reconstructed 4D scene and the user’s keyboard commands to direct the next round of video generation. The generation and reconstruction steps proceed in real time, with only minimal latency between guidance and generation. This cyclic process continuously updates a 4D spatial–temporal memory of the constructed dynamic scene, allowing effective motion and interaction to be driven interactively via keyboard control.

\subsection{Long-memory Auto-regressive Video Generation}
\subsubsection{Micro and Macro Planning}
\label{method_sec_1}
Motivated by the analysis in the MMPL~\cite{xiang2025MMPL}, we observe that 
autoregressive models accumulate errors proportionally to the number of propagation steps, 
whereas non-autoregressive models decouple errors from the step count through joint optimization. 
To exploit the complementary strengths of both paradigms, 
we introduce \textit{Macro-from-Micro Planning (MMPL)} into our TeleWorld, 
a unified planning method comprising two key components: 
\textit{Micro-Planning} and \textit{Macro-Planning}.
\label{train_method}

\textbf{Micro Planning.}
\label{Micro}
Micro Planning $\mathcal{M}_s$ builds a short-term narrative for the $s$-th segment by predicting a sparse set of key frames $\mathcal{P}_{\mathcal{M}_s} = \{x_s^{t_a}, x_s^{t_b}, x_s^{t_c}\}$ from the initial frame $x_s^1$. These \textit{pre-planning frames} serve as stable anchors for subsequent synthesis, with timestamps set as $t_a = 2$ (early neighbor), $t_b = N/2$ (midpoint), and $t_c = N$ (segment end).  
The process is formulated as:  
\begin{equation}
p(\mathcal{P}_{\mathcal{M}_s} \mid x_s^1)
= p(x_s^{t_a}, x_s^{t_b}, x_s^{t_c} \mid x_s^1).
\end{equation}

All frames are jointly optimized conditioned only on $x_s^1$, which mutually constrains their residual errors and eliminates cumulative drift—unlike sequential autoregressive generation. This design ensures within-segment coherence and provides a drift-resistant foundation for later content population.

\begin{figure*}[t]
    \includegraphics[width=\textwidth]{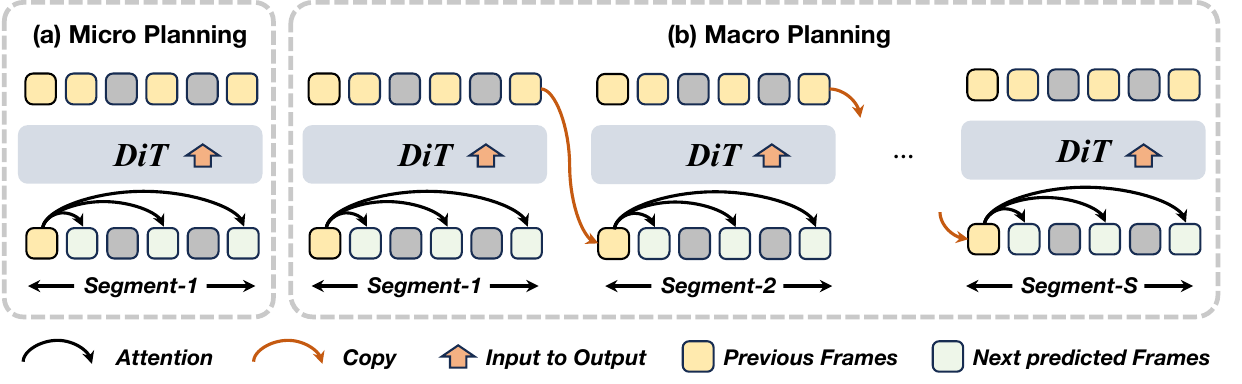}
    \vspace{-0.05in}
    \caption{Our macro-from-micro planning framework is organized into two levels: (1) Micro Planning, where a sequence of frames is generated within each local segment to constrain error propagation; and (2) Macro Planning, which links segments through an autoregressive chain—each step’s output frames guide the prediction of the next, ensuring long-range temporal consistency.
    As shown in the figure, the three predicted frames marked in green correspond to the initial pre-planning frames, $\mathcal{P}_{\mathcal{M}_s} = \{x_s^{t_a}, x_s^{t_b}, x_s^{t_c}\}$ , which serve as keyframes to maintain long-term memory and stability throughout the video sequence.}
    \label{fig:ours}
    \vspace{-0.1in}
\end{figure*}

\textbf{Macro Planning.}
\label{Macro}
While Micro Planning provides a segment-level temporal storyline,
it remains limited in capturing global dependencies across the entire videos of the world scene. To achieve long-range coherence, we extend it into \textbf{Macro Planning}, denoted \(\mathcal{M}^{+}\). This constructs a global storyline by chaining overlapping Micro Plans sequentially across segments: the terminal pre-planning frame of one segment initializes the next, forming a segment-level autoregressive chain along the video timeline.

Formally, given a full video of length \(T\) partitioned into \(S\) segments, let \(x_s^1\) be the initial frame of the \(s\)-th segment. The set of planning frames produced by Macro Planning is denoted \(\mathcal{P}_{\mathcal{M}^{+}}\). The process is defined as:
\begin{equation}
p(\mathcal{P}_{\mathcal{M}^{+}} \mid x_1^1)
= \prod_{s=1}^{S} p(\mathcal{P}_{\mathcal{M}_s} \mid x_s^1),
\quad
x_{s+1}^1 := x_s^{t_c},
\quad
\mathcal{P}_{\mathcal{M}^{+}} := \bigcup_{s=1}^{S} \mathcal{P}_{\mathcal{M}_s}.
\end{equation}
Here, \(\mathcal{M}_s\) is the Micro Planning for segment \(s\). By linking segments hierarchically, Macro Planning converts frame-by-frame autoregressive dependencies into a sparse sequence of segment-level planning steps. This ensures consistent global narrative flow, mitigates temporal drift, and reduces error accumulation from the scale of \(T\) frames to only \(S\) segments, where \(S \ll T\).

This hierarchical linking enables the world model to retain long-term memory across segments. Subsequently, we anchor these memories through a online 4D reconstruction of the cross-segment anchor frames, embedding all keyframes within a coherent spatio-temporal field. This further clarifies and stabilizes the inter-segment and intra-segment memory, ensuring its precision and consistency.

However, when chaining Micro Plannings autoregressively, directly using the tail latent tokens of one segment as the prefix for the next often introduces boundary flickering and color shifts due to distribution mismatch between initial and temporally-compressed latent frames.

To stabilize inter-segment transitions, we adopt a drift-resilient re-encoding and decoding strategy. Specifically, we reconstruct a short video clip from the concatenated initial and terminal planning tokens of the current segment. To ensure temporal continuity during decoding, the terminal tokens are duplicated and inserted to form a contiguous latent sequence. The re-encoded latents of the second copy then serve as the initial condition for the next segment. For implementation details, we refer readers to our previous work MMPL~\cite{xiang2025MMPL}.

\subsubsection{MMPL-based Content Populating}
\label{Populating}
Following Sec.~\ref{method_sec_1}, the Micro Plan $\mathcal{M}_s$ divides each video segment into two sub-segments—e.g., $\big[x_s^{t_a}, x_s^{t_b}\big]$ and $\big[x_s^{t_b}, x_s^{t_c}\big]$—bounded by consecutive planning frames. To synthesize the full segment by filling the remaining frames under the guidance of these planning anchors, we introduce MMPL-based Content Populating.

Micro Planning provides three types of key frames: \textit{early} ($x_s^{t_a}$), \textit{midpoint} ($x_s^{t_b}$), and \textit{terminal} ($x_s^{t_c}$). Motivated by earlier frame-conditioned generation approaches, we perform content population in two sequential stages:
\begin{enumerate}
    \item Populate the first sub-segment using the initial frame and the early planning frame as the start, and the midpoint planning frame as the end.
    \item Extend the sequence by taking all frames up to the midpoint as the new start and the terminal frame as the end, thereby generating the remaining content.
\end{enumerate}
The process can be formally expressed as:
\begin{equation}
\label{eq:pop}
p(\mathcal{C}_s \mid \mathcal{P}_{\mathcal{M}_s})
= p\big(x_s^{t_a+1:t_b-1} \mid x_s^{1:t_a}, x_s^{t_b}\big)
\cdot
p\big(x_s^{t_b+1:t_c-1} \mid x_s^{1:t_b}, x_s^{t_c}\big),
\end{equation}
Here, $\mathcal{C}_s$ denotes the content frames to be generated in segment $s$, while $x_s^{t_a}$, $x_s^{t_b}$, and $x_s^{t_c}$ represent its early, midpoint, and terminal planning frames, respectively. The notation $x_s^{1:t_a}$ and $x_s^{1:t_b}$ indicates that the generation of each sub-segment is conditioned on all preceding frames within the segment, in addition to its boundary planning frames. The intermediate frames $x_s^{t_a+1:t_b-1}$ and $x_s^{t_b+1:t_c-1}$ correspond to the content to be populated.

Importantly, the factorization in Eq.~\ref{eq:pop} shows that content population within each sub-segment depends solely on its corresponding planning frames. This allows multiple sub-segments to be optimized in parallel once their internal planning frames are ready. By distributing segment-wise optimization across multiple GPUs, the proposed MMPL-based Content Populating enables concurrent execution, significantly accelerating the synthesis of long videos.

\subsection{Real-time 4D Reconstruction}
\subsubsection{Key-frame Reconstruction}
As discussed in Introduction, we propose a real-time 4D reconstruction module to further provide dynamic memory of moving objects within the scene. Considering the planning strategy in the MMPL architecture, our reconstruction process also follows macro planning synchronously. The reconstruction task continuously progresses backward along with the macro structure, allowing the reconstruction speed to closely follow the generation process. Meanwhile, micro planning uses the rendered results of the reconstruction under corresponding manipulations as guidance. 

In this way, the overhead of reconstruction is minimized, and the input to reconstruction is kept as sparse as possible to prevent the reconstruction task from failing over long sequences due to extended world generation. We term this approach key-frame reconstruction.

Specifically, only the sparse set of \textit{pre-planning frames} $\mathcal{P}_{\mathcal{M}_s} = \{x_s^{t_a}, x_s^{t_b}, x_s^{t_c}\}$  need to conduct 4D reconstruction. These planning frames essentially serve as anchors within the video—they are generated first with minimal error and highest quality, and they determine the motion trajectory of the video. Using them for 4D reconstruction also introduces sufficiently rich records for long‑video generation tasks in world models. The beginning, middle, and end of each video segment will be used to record information in the 4D spatiotemporal field. During content population, the intermediate motion is then filled in based on these recorded cues.

\subsubsection{Move Obejct Segmentation}
Inspired by 4D-VGGT~\cite{wang20254dvggtgeneralfoundationmodel}, we utilize its dynamic saliency map as the dynamic masks. To aggregate temporal information, we employ an interframe sliding-window strategy across frames, defined as $\mathcal{W}(t)=\{t-n, \ldots, t-1, t+1, \ldots, t+n\}$. Within this window and across three set of layers $L$, including shallow, middle, and deep layers
Shallow, middle, and deep correspond to different layer ranges \((i,j)\). \(w_{\mathrm{shallow}}\) captures semantic saliency, \(w_{\mathrm{middle}}\) reflects motion instability, and \(w_{\mathrm{deep}}\) provides a spatial prior to suppress outliers. Finally, a per-frame dynamic mask is obtained by thresholding: \(M_t = [\mathrm{Dyn} > \alpha]\), followed by feature clustering for refinement. A network-level early-stage masking strategy for 4D reconstruction and stacking is also conducted in our framework. Static scene elements are merged and progressively expanded, while sparse dynamic components are separately rendered over time. However, since our input is limited to \textit{pre-planning frames} $\mathcal{P}_{\mathcal{M}_s} = \{x_s^{t_a}, x_s^{t_b}, x_s^{t_c}\}$ , the rendered dynamic content remains highly sparse. This requires predicting subsequent dynamic regions based on earlier frames within the pre-planning sequence—a challenge we address through macro-planning in video generation. From a macroscopic perspective, smooth continuous motion is decomposed into keyframe-like dynamic segments embedded within the scene.

Specifically, following 4D-VGGT~\cite{wang20254dvggtgeneralfoundationmodel}, to mitigate geometric inconsistencies introduced by dynamic pixels, we also mask dynamic image tokens only in shallow and mid-level layers (layers 1$\sim$5) by suppressing their Key (K) vectors.


\subsection{Guidance}
\subsubsection{Keyboard Control}
As the widespread adoption of keyboard control in world models~\cite{mao2025yume,tang2025hunyuangamecraft2instructionfollowinginteractivegame,hong2025relicinteractivevideoworld}, we also utilize the four WASD keys along with the arrow keys to simulate movement and perspective changes, as illustrated below. These inputs are correspondingly mapped to camera poses. 

These signals are conditioned to guide the model's generation. We map these controls into camera motion movements along with input frame depth scales.

\noindent
\begin{minipage}[t]{0.48\textwidth}
\centering
\scriptsize
$
\begin{aligned}
& \begin{array}{l}
    \text{perspective} \\
    \text{changes}
   \end{array} =
\left\{
\begin{array}{l}
\rightarrow \text{ : Camera turns right }(\rightarrow). \\
\leftarrow \text{ : Camera turns left }(\leftarrow). \\
\uparrow \text{ : Camera tilts up }(\uparrow). \\
\downarrow \text{ : Camera tilts down }(\downarrow). \\
\uparrow\rightarrow \text{ : Camera tilts up and turns right }(\uparrow\rightarrow). \\
\downarrow\rightarrow \text{ : Camera tilts down and turns right }(\downarrow\rightarrow). \\
\downarrow\leftarrow \text{ : Camera tilts down and turns left }(\downarrow\leftarrow). \\
\cdot \text{ : Camera remains still }(\cdot).
\end{array}
\right. \\
\end{aligned}
$
\end{minipage}
\hfill
\begin{minipage}[t]{0.48\textwidth}
\centering
\scriptsize
$
\begin{aligned}
& \begin{array}{l}
    \text{camera} \\
    \text{movement}
   \end{array} =
\left\{
\begin{array}{l}
\text{W : Camera moves forward (W).} \\
\text{A : Camera moves left (A).} \\
\text{S : Camera moves backward (S).} \\
\text{D : Camera moves right (D).} \\
\text{W+A : Camera moves forward and left (W+A).} \\
\text{W+D : Camera moves forward and right (W+D).} \\
\text{S+D : Camera moves backward and right (S+D).} \\
\text{S+A : Camera moves backward and left (S+A).} \\
\text{None : Camera stands still }(\cdot).
\end{array}
\right. \\
\end{aligned}
$
\end{minipage}

Furthermore, to enhance the continuity and coherence of video generation as much as possible, we endeavor to avoid maintaining a static camera position. Therefore, even when no keyboard input is provided by the user, the camera pose will drift forward at a very slow speed—a feature we refer to as the standby animation.

\subsubsection{View-Conditioned Guidance}
Subsequently, we need to encode the processed keyboard inputs for the world model network. As discussed in ReCamMaster~\cite{}, the conditioning by frame dimension is a more effective approach for integrating target camera poses into the DiT network. Following this insight, we adopt a similar structure and incorporate the following mechanism into TeleWorld's DiT network:

To achieve better synchronization and content consistency with the keyboard guidance video, we propose to concatenate the guidance video tokens with the target video tokens along the frame dimension:

$$
\left\{
\begin{array}{l}
x_s=\text{patchify}\left(z_s\right), \quad x_t=\text{patchify}\left(z_t\right), \\
x_i=\left[x_s, x_t\right]_{\text{frame-dim}},
\end{array}
\right.
$$

where $x_i \in \mathbb{R}^{b \times 2f \times s \times d}$ is the input of the diffusion transformer. In other words, the input token number is doubled compared to the vanilla video generation process. Moreover, no additional attention layers are needed for cross-video aggregation, as 3D self-attention inherently processes all tokens.


\subsection{Distribution Matching Distillation}
Our approach integrates seamlessly with existing Distribution Matching Distillation (DMD) frameworks without requiring any architectural modifications. Specifically, the MMPL video generation pipeline adjusts the attention visibility range and prediction order during both training and inference. Building on standard self-forcing pipelines, DMD can be directly applied on top of MMPL and deployed within the TeleWorld framework.

When combined with parallelized decoding, the resulting system delivers substantial inference speedups, achieving sustained throughput exceeding 32 FPS for long-horizon video generation on the TeleWorld-1.3B model and 8 FPS on the Teleworld-18B model, both evaluated on NVIDIA H100 GPUs.

Despite its importance for real-time video generation, DMD introduces significant challenges to the training infrastructure, esp. when applied to our 18B model. The training setup requires the simultaneous coordination of three diffusion models—the autoregressive generator, the critic, and the teacher—making it infeasible to host all components within a single 80-GB HBM GPU. To address this constraint, we employ Ray, \cite{PhilippRay}, to distribute the model weights across multiple GPUs. Furthermore, leveraging the Ulysses sequence-parallel capabilities provided by TeleTron\footnote{https://github.com/Tele-AI/TeleTron}
, we shard the generator’s KV cache across GPUs, enabling it to fit within memory limits.

\begin{figure}[t]
    \centering

    \begin{subfigure}[t]{0.6\linewidth}
        \centering
        \includegraphics[width=\linewidth, trim=0 0 0 0]{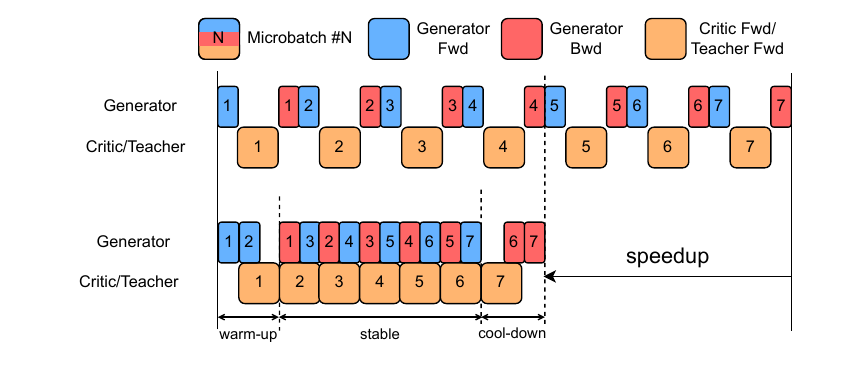}
        \caption{Pipeline execution schedule for generator steps. }
        \label{fig:generator-schedule}
    \end{subfigure}
    \hfill
    \begin{subfigure}[t]{0.37\linewidth}
        \centering
        \includegraphics[width=\linewidth, trim=10 10 10 5]{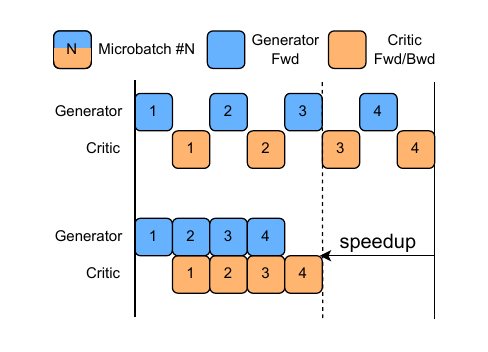}
        \caption{Pipeline execution schedule for critic steps.}
        \label{fig:critic-schedule}
    \end{subfigure}

    \caption{ Pipeline execution schedules for Distribution-Matching Distillation.
(a) Generator-step pipeline with 7 micro-batches. Cell length denotes execution time. The critic and teacher works in parallel, so their cells are merged together for simplicity, and their cell length denotes the maximum of their execution time. The upper half of the figure is the non-pipelining baseline, which introduces a lot of GPU bubbles (i.e. GPU idle time). The lower half is our proposed pipeline schedule. In the stable phase, the generator backward stage of micro-batch $i$ and the generator forward stage of micro-batch $i+2$ are executed concurrently with the critic/teacher forward stage of micro-batch $i+1$. The execution time of all stages are carefully balanced by allocating appropriate numbers of GPUs to each component, enabling near-perfect overlap. This method minimizes GPU bubbles and achieves efficient parallelization of generator, teacher, and critic workloads in the proposed system.
(b) Critic-step pipeline with 4 micro-batches. Since the generator parameters remain frozen during the critic update, the pipeline follows a simpler producer-consumer execution pattern.}
    \label{fig:pipeline-schedule}
\end{figure}

To mitigate GPU underutilization caused by model parallelism, we design a novel \textbf{pipelined training schedule} that overlaps the computation of the generator, critic, and teacher models, thereby minimizing GPU idle time (i.e., pipeline bubbles). The execution schedules for the generator and critic steps are illustrated in Figure~\ref{fig:pipeline-schedule}. For the generator step, enabling the degree of overlap shown in the figure requires carefully matching the combined execution time of the generator forward and backward stages to that of the critic/teacher stage through explicit resource allocation. In practice, we find that a generator:critic:teacher GPU ratio of 4:1:1 achieves near-perfect overlap. In addition, to simplify DMD optimization and ensure predictable stage durations, we fix the number of denoising steps in the generator pipeline during training rather than randomly sampling them. We note that two copies of the KV cache must be maintained to support correct backpropagation; however, this overhead is manageable since the KV cache is already sharded across devices using context parallelism. As a result, our pipelined system achieves an approximately 50\% end-to-end training speedup compared to a non-pipelined baseline. 

Taken together, efficient KV-cache sharding, model parallelism, and pipelined execution position our training system to scale naturally to future auto-regressive diffusion models with substantially larger parameter counts.

\subsection{Streaming and Scheduled Generation with Online Video Super-resolution}

\begin{figure}[t]
    \centering
    \includegraphics[width=\textwidth]{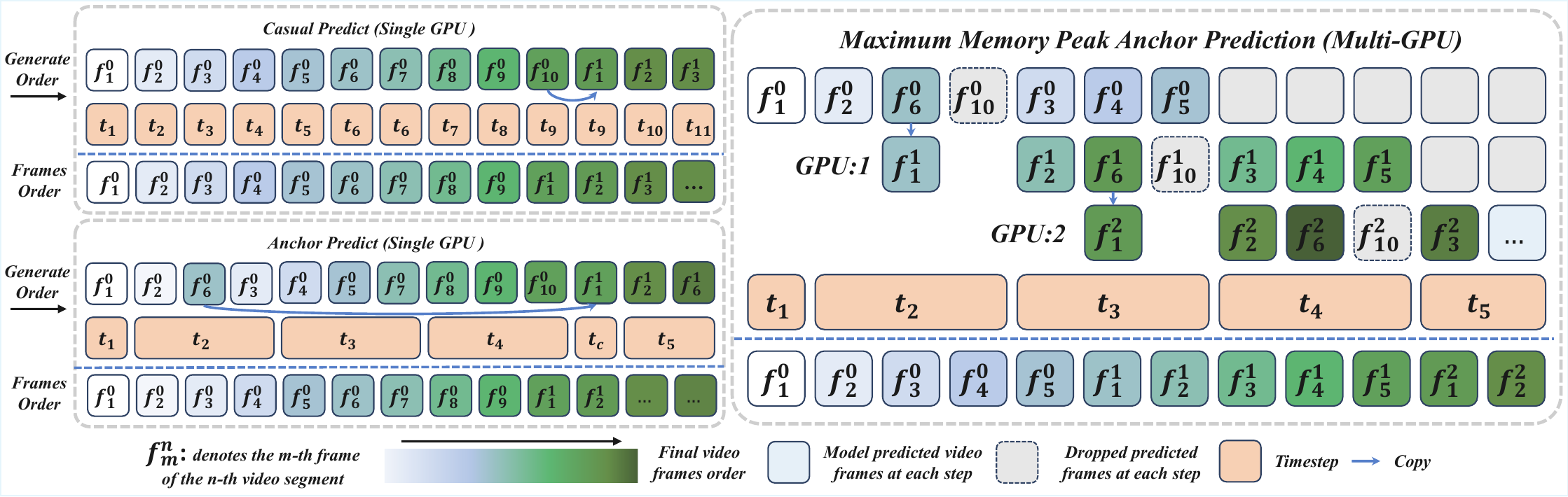}
    \vspace{-0.2in}
    \caption{
    Multi-GPU parallel inference via adaptive workload scheduling.  
    Given the initial frame $f^0_1$, segment 0 first generates its planning frames  
    $f^0_2$, $f^0_6$, and $f^0_{10}$.  
    These planning frames then guide the content population of the intermediate frames  
    $f^0_3$, $f^0_4$, and $f^0_5$.  
    While segment 0 is still populating these frames,  
    segment 1 can immediately start its Micro Planning  
    by taking $f^0_{10}$ as the initial frame $f^1_1$  
    and generating its own planning frames $f^1_2$, $f^1_6$, and $f^1_{10}$.  
    This staged execution enables overlapping planning and populating across segments,  
    maximizing multi-GPU parallelism. Here, each $t_i$ denotes an inference step in the diffusion sampling process.
    }
    \label{fig:parallel_inference_method}
    \vspace{-0.1in}
\end{figure}
\subsubsection{Scheduled Generation}
Although content populating across different segments can be parallelized (Sec.~\ref{Populating}), a key limitation remains: parallel execution cannot begin until planning frames for all segments are fully generated, leading to an unavoidable prefix delay that reduces overall throughput.

To address this, we introduce an \textit{adaptive workload scheduling} strategy that dynamically orders the execution of Micro Planning, Macro Planning, and Content Populating to maximize parallelism. Since Macro Planning forms an autoregressive chain of segment-level Micro Plannings, the planning frames are generated sequentially across segments. This allows the Content Populating of an earlier segment to start as soon as its own planning frames are ready, without waiting for subsequent segments.

For illustration, with $t_a=2$, $t_b=6$, and $t_c=10$, the planning frame $x^{t_c}_s$ from the current segment immediately serves as the initial frame $x^1_{s+1}$ for the next segment. Thus, the next segment can begin its Micro Planning while the current one is still populating its intermediate frames (e.g., $x^{t_a+1:t_b-1}_s$).
This staged independence naturally enables segment-parallel generation, 
as formally expressed in Eq.~(\ref{eq:segment_parallel}):
\begin{equation}
\begin{aligned}
& \text{Segment s:} && x_{s}^{t_a+1:t_b-1} 
   \sim p_\theta(x \mid x_{s}^{1}, x_{s}^{t_a}, x_{s}^{t_b}), \\
& \text{Segment s+1:} && \{x_{s+1}^{t_a}, x_{s+1}^{t_b}, x_{s+1}^{t_c}\} 
   \sim p_\theta(x \mid x^1_{s+1}), 
   \quad x^1_{s+1} \in \{x_{s}^{t_b}, x_{s}^{t_c}\}.
\end{aligned}
\label{eq:segment_parallel}
\end{equation}
Here, the initial frame \(x^1_{s+1}\) of the next segment can be selected 
either as \(x_{s}^{t_b}\) or \(x_{s}^{t_c}\). 
In order to keep the real-time practical generation, we choose the maximum throughput prediction as follows:

To minimize latency as much as possible, we use the \textbf{Minimum Memory Peak Prediction} strategy.
When $x^{t_b}_s$ is used as $x^1_{\text{s+1}}$, 
intermediate frames $x^{t_b+1}:x^{t_c-1}$ are skipped, 
bypassing the region with the deepest temporal context 
and highest generation latency. 
This mode minimizes peak memory usage and reduces per-segment latency 
but introduces frame reuse between segments, 
slightly reducing overall throughput. As illustrated in Fig. \ref{eq:segment_parallel}, \( f_4^0 \) and \( f_6^1 \) are in fact generated synchronously. This means that any immediate user input manipulation is only rendered after three latent chunks, resulting in a feedback latency of approximately one second. Consequently, the world output currently being observed corresponds to the pre-buffered changes captured one second prior to the user’s input.


\subsubsection{Streamed VAE}
To achieve real-time video generation for live streaming, we designed a streaming-capable VAE based on the principles of StreamDiffusionV2~\cite{feng2025streamdiffusionv2streamingdynamicinteractive}. The core challenge in a live setting is to minimize the “time to first frame” and ensure continuous, low-latency output, which is fundamentally different from batch-based video generation that processes long sequences offline. Our Stream-VAE is a low-latency Video-VAE variant specifically optimized for streaming inference. Instead of encoding an entire video sequence at once, it operates on short, contiguous video chunks—typically 4 frames in our implementation. This chunk-wise processing is critical for maintaining a steady output stream.

The architecture of the Stream-VAE incorporates strategic caching of intermediate features within its 3D convolutional layers. As each new chunk of frames is fed into the model, the network reuses relevant temporal features computed from previous chunks, thereby preserving temporal coherence across chunk boundaries without the need to re-encode a long history. This design significantly reduces redundant computation and memory overhead, enabling efficient incremental encoding and decoding. By integrating this Stream-VAE into our pipeline, we ensure that the latent representations of the video are generated and can be delivered to users with minimal delay, forming the foundational stage of our real-time streaming system.

\subsubsection{Video Super-resolution}
For the subsequent enhancement of video quality, we incorporate a streaming super-resolution module inspired by FlashVSR~\cite{}. This component is responsible for upscaling the decoded latents from the Stream-VAE into high-resolution video frames in real time. A key innovation we adopt from FlashVSR is its locality-constrained sparse attention mechanism. This mechanism restricts the self-attention operations to local spatial-temporal windows, drastically reducing the computational complexity that typically plagues video super-resolution models. It effectively bridges the resolution gap often encountered between training and inference without sacrificing the quality of fine details.

Furthermore, we leverage FlashVSR's lightweight conditional decoder, which is engineered for fast feature reconstruction. The decoder conditions its upscaling process on the features extracted from the Stream-VAE's output, ensuring high-fidelity results while maintaining a low computational footprint. Crucially, this super-resolution module is designed to work in harmony with our Stream-VAE in a fully streaming manner. It processes short video chunks (e.g., 5 frames) that align with the VAE's output stream, applying super-resolution incrementally as each chunk becomes available. This integrated, chunk‑wise processing pipeline enables our model to achieve super‑resolution decoding at approximately 17 FPS on 960$\times$1760 resolution videos, making high‑quality real‑time video generation practical.

In summary, by integrating Scheduled Generation, Streamed VAE, and Video Super‑resolution techniques, our system enables the TeleWorld‑18B model to achieve stable 8 FPS performance and generate high‑quality 960$\times$1760 videos on a setup of four NVIDIA H100 GPUs.

\section{Experiments and Discussion}

\subsection{Multi-modal Dataset Preparation}
We introduce the data collections here.
To support large-scale training and unified evaluation, we construct TeleWorld-500K, a curated dataset tailored for controllable camera and dynamics 4D annotated videos. TeleWorld-500K is built through two pipelines.

\subsubsection{Curation Pipeline}
\noindent\textbf{(1) Data Collection.}
We assembled a large-scale collection of real-world video clips through a hybrid approach combining systematic web scraping and selective manual gathering. Sources included major public platforms such as YouTube, Pexels, Pixabay, Mixkit, and Bilibili, ensuring broad coverage of diverse visual content and scenarios.

\noindent\textbf{(2) Automated Quality Filtering.}
From the initial pool, we applied a multi-stage automated filtering pipeline to eliminate low-quality content. The LAION aesthetic scorer was used to retain clips with aesthetic ratings above 6, while PaddleOCR~\cite{liao2022real} detected and removed videos containing prominent overlaid text, watermarks, or subtitles. Additionally, extremely short, corrupted, or visually inconsistent clips were automatically discarded to maintain overall dataset integrity.

\noindent\textbf{(3) Motion-Aware Selection.}
To ensure the dataset contains meaningful dynamics suitable for controllable camera and object modeling, we performed motion-based filtering. Using TTT3R~\cite{chen2025ttt3r}, we estimated per-clip camera motion and excluded sequences with negligible viewpoint changes. Furthermore, to retain videos with salient foreground object motion, the vision-language model Qwen-2.5-VL-72B~\cite{bai2025qwen2} analyzed each clip and filtered out those without detectable moving subjects.

\noindent\textbf{(4) Expert Review and Dataset Finalization.}
The remaining clips underwent thorough manual inspection by twenty domain experts over 690 person-hours to remove any residual low-quality or unsuitable content. This careful curation resulted in the final TeleWorld-500K dataset, comprising 500K high-quality video clips that feature diverse real-world environments, pronounced camera motion, and rich dynamic interactions, providing a robust foundation for training world models.

\subsubsection{Annotation Pipeline}

\noindent\textbf{(1) Motion Object Segmentation.}
To annotate moving objects, we first employed Segment Any Motion in Videos~\cite{huang2025segment}, which takes a video as input and predicts masks for all moving foreground objects. It provides an initial mask on the first frame for each distinct object, with unique colors assigned to maintain consistent identity labeling across frames.

\noindent\textbf{(2) Camera Trajectory Annotating.}
Using the first-frame object masks as initialization, we employed 4D-VGGT~\cite{wang20254dvggtgeneralfoundationmodel} to recover dense motion and camera annotations. 4D-VGGT is a unified camera trajectory annotating framework that jointly estimates point clouds, depth maps, camera intrinsics, and camera poses in an end-to-end manner. For each video, it reconstructs 3D trajectories of moving objects and estimates per-frame camera poses.

\noindent\textbf{(3) Semantic Description Generation.} To enable precise text description, we employed the large vision–language model Qwen-2.5-VL-72B~\cite{bai2025qwen2} to generate textual annotations that describe the appearance and motion of both moving object and camera motion, along with the overall scene context. These captions complement the 3D trajectories of moving objects, providing comprehensive semantic information aligned with scene-level dynamics.


\subsection{WorldScore Benchmark}
This section evaluates TeleWorld on the WorldScore~\cite{duan2025worldscoreunifiedevaluationbenchmark} benchmark, which is currently one of the most comprehensive protocols for measuring ``world generation'' ability. Unlike image or short-video benchmarks that primarily assess local visual quality, WorldScore evaluates whether a model can construct and maintain a consistent world across viewpoints, scene transitions, and temporal evolution. The benchmark includes both static and dynamic settings, as well as a rich set of metrics assessing controllability, consistency, perceptual quality, and motion behavior. All results in this section are reported from the official WorldScore leaderboard to ensure comparability.

The WorldScore evaluation consists of two primary aggregate dimensions. First, WorldScore-Static measures whether the generated world remains stable and coherent while the camera moves through multiple viewpoints. This focuses on spatial fidelity, layout preservation, and cross-view semantic consistency. Second, WorldScore-Dynamic measures world evolution over time, including object motion, scene changes, and temporal stability. This dimension evaluates whether a model generates motion patterns that are coherent, semantically grounded, and structurally consistent with the underlying world. The official evaluation pipeline computes a set of sub-metrics and integrates them into the two final aggregate scores.

WorldScore reports 12 metrics. Camera Control, Object Control, and Content Alignment measure controllability. They jointly characterize how well a model follows layout constraints, preserves required entities, and responds to semantic instructions. 3D Consistency, Photometric Consistency, Style Consistency, and Subjective Quality measure structural and perceptual stability. These metrics reflect how well a model maintains consistent geometry, appearance, lighting, and aesthetics. Motion Accuracy, Motion Magnitude, and Motion Smoothness measure dynamic behavior, capturing temporal realism, motion amplitude suitability, and continuity. Together, these metrics serve as a comprehensive evaluation of static world structure and dynamic world evolution.

\begin{table}[htbp]
\centering
\scriptsize
\setlength{\tabcolsep}{3pt}
\renewcommand{\arraystretch}{1.05}
\resizebox{\textwidth}{!}{%
\begin{tabular}{lccccccccc}
\hline
Model Name & WS-Static & WS-Dynamic & CamCtrl & ObjCtrl & ContAlign & 3DCons & PhotoCons & StyleCons & SubjQual \\
\hline
TeleWorld & \textbf{78.23} & \textbf{66.73} & 76.58 & \textbf{74.44} & 73.20 & 87.35 & 88.82 & \textbf{85.59} & 61.66 \\
Voyager~\cite{Voyager} & 77.62 & 54.53 & 85.95 & 66.92 & 68.92 & 81.56 & 85.99 & 84.89 & 71.09 \\
WonderWorld~\cite{Yu_2025_Wonderworld} & 72.69 & 50.88 & 92.98 & 51.76 & 71.25 & 86.87 & 85.56 & 70.57 & 49.81 \\
LucidDreamer~\cite{chung2023luciddreamer} & 70.40 & 49.28 & 88.93 & 41.18 & 75.00 & 90.37 & 90.20 & 48.10 & 58.99 \\
WonderJourney~\cite{yu2023wonderjourney} & 63.75 & 44.63 & 84.60 & 37.10 & 35.54 & 80.60 & 79.03 & 62.82 & 66.56 \\
CogVideoX-I2V~\cite{yang2025cogvideoxtexttovideodiffusionmodels} & 62.15 & 59.12 & 38.27 & 40.07 & 36.73 & 86.21 & 88.12 & 83.22 & 62.44 \\
Text2Room~\cite{Hollein_2023_text2room} & 62.10 & 43.47 & 94.01 & 38.93 & 50.79 & 88.71 & 88.36 & 37.23 & 36.69 \\
InvisibleStitch~\cite{InvisibleStitch} & 61.12 & 42.78 & 93.20 & 36.51 & 29.53 & 88.51 & 89.19 & 32.37 & 58.50 \\
Gen-3~\cite{gen3} & 60.71 & 57.58 & 29.47 & 62.92 & 50.49 & 68.31 & 87.09 & 62.82 & 63.85 \\
Wan2.1~\cite{Wan} & 57.56 & 52.85 & 23.53 & 40.32 & 45.44 & 78.74 & 78.36 & 77.18 & 59.38 \\
Hailuo~\cite{minimax} & 57.55 & 56.36 & 22.39 & 69.56 & 73.53 & 67.18 & 62.82 & 54.91 & 52.44 \\
LTX-Video~\cite{hacohen2024ltx} & 55.44 & 56.54 & 25.06 & 53.41 & 39.73 & 78.41 & 88.92 & 53.50 & 49.08 \\
Allegro~\cite{zhou2024allegro} & 55.31 & 51.97 & 24.84 & 57.47 & 51.48 & 70.50 & 69.89 & 65.60 & 47.41 \\
CogVideoX-T2V~\cite{yang2025cogvideoxtexttovideodiffusionmodels} & 54.18 & 48.79 & 40.22 & 51.05 & 68.12 & 68.81 & 64.20 & 42.19 & 44.67 \\
EasyAnimate~\cite{xu2024beasyanimate} & 52.85 & 51.65 & 26.72 & 54.50 & 50.76 & 67.29 & 47.35 & 73.05 & 50.31 \\
VideoCrafter2~\cite{chen2023videocrafter1} & 52.57 & 47.49 & 28.92 & 39.07 & 72.46 & 65.14 & 61.85 & 43.79 & 56.74 \\
DynamiCrafter~\cite{xing2023dynamicrafter}  & 52.09 & 47.19 & 25.15 & 47.36 & 25.00 & 72.90 & 60.95 & 78.85 & 54.40 \\
SceneScape~\cite{fridman2024scenescape} & 50.73 & 35.51 & 84.99 & 47.44 & 28.64 & 76.54 & 62.88 & 21.85 & 32.75 \\
VideoCrafter1-I2V~\cite{chen2023videocrafter1} & 50.47 & 47.64 & 25.46 & 24.25 & 35.27 & 74.42 & 73.89 & 65.17 & 54.85 \\
VideoCrafter1-T2V~\cite{chen2023videocrafter1} & 47.10 & 43.54 & 21.61 & 50.44 & 60.78 & 64.86 & 51.36 & 38.05 & 42.63 \\
T2V-Turbo~\cite{li2024t2v} & 45.65 & 40.20 & 27.80 & 30.68 & 69.14 & 38.72 & 34.84 & 49.65 & 68.74 \\
Vchitect-2.0~\cite{fan2025vchitect}  & 42.28 & 38.47 & 26.55 & 49.54 & 65.75 & 41.53 & 42.30 & 25.69 & 44.58 \\
4D-fy~\cite{bahmani20244dfy} & 27.98 & 32.10 & 69.92 & 55.09 & 0.85 & 35.47 & 1.59 & 32.04 & 0.89 \\
\hline
\end{tabular}%
}
\caption{Quantitative comparison on the WorldScore benchmark. We report the leaderboard scores for static and dynamic world generation (WorldScore-Static/Dynamic) and the corresponding controllability and consistency metrics (Camera Control, Object Control, Content Alignment, 3D/Photometric/Style Consistency, and Subjective Quality) for TeleWorld and representative baselines under the official evaluation protocol. Higher is better for all metrics.}
\label{tab:worldscore_core_metrics}
\end{table}

\subsubsection{Quantitative Results: }
We compare TeleWorld against 23 baseline models across 3D, 4D, and video-based approaches. These baselines include 3D world generators such as Voyager~\cite{Voyager}, WonderWorld~\cite{Yu_2025_Wonderworld}, LucidDreamer~\cite{chung2023luciddreamer}, WonderJourney~\cite{yu2023wonderjourney}, Text2Room~\cite{Hollein_2023_text2room}, InvisibleStitch~\cite{InvisibleStitch}, and SceneScape~\cite{fridman2024scenescape}; 4D-oriented systems such as 4D-fy~\cite{bahmani20244dfy}; and a range of image-to-video and text-to-video systems including Gen-3~\cite{gen3}, Wan2.1~\cite{Wan}, Hailuo~\cite{minimax}, LTX-Video~\cite{hacohen2024ltx}, Allegro~\cite{zhou2024allegro}, CogVideoX~\cite{yang2025cogvideoxtexttovideodiffusionmodels}, EasyAnimate~\cite{xu2024beasyanimate}, DynamiCrafter~\cite{xing2023dynamicrafter}, VideoCrafter~\cite{chen2023videocrafter1}, T2V-Turbo~\cite{li2024t2v}, and Vchitect~\cite{fan2025vchitect}. All compared models are evaluated under the same protocol. TeleWorld is tested under the Video and I2V configuration using a single generation setup not specialized for the WorldScore benchmark.

TeleWorld achieves the strongest performance on both aggregate metrics, with a WorldScore-Static score of 78.23 and a WorldScore-Dynamic score of 66.73. The next best models achieve 77.62 in the static setting (Voyager~\cite{Voyager}) and 59.12 in the dynamic setting (CogVideoX-I2V~\cite{yang2025cogvideoxtexttovideodiffusionmodels}). TeleWorld therefore outperforms the strongest baselines by 0.61 points in static world generation and by 7.61 points in dynamic world generation. The relatively small margin in static performance indicates that TeleWorld reaches the saturation point of current static scene modeling, while the significantly larger dynamic margin suggests a distinct advantage in temporal reasoning, motion modeling, and evolving world stability. Notably, TeleWorld is the only method that simultaneously ranks first in both static and dynamic tracks, indicating that it does not favor one operational regime at the cost of the other.

In controllability, TeleWorld delivers balanced scores in Camera Control (76.58), Object Control (74.44, best among all systems), and Content Alignment (73.20). This indicates that it respects multi-modal user constraints without specializing in a single dimension. The strong Object Control score, in particular, suggests that TeleWorld maintains an implicit, persistent world state that preserves object identity and arrangement across long sequences, consistent with its closed-loop generation–reconstruction design.

TeleWorld also excels in structural and perceptual consistency, with scores of 87.35 (3D Consistency), 88.82 (Photometric Consistency), 85.59 (Style Consistency), and 61.66 (Subjective Quality). These results reflect that the generated content behaves as projections of a coherent internal 4D representation—aligned with our framework’s ability to capture and enforce global spatio-temporal structure while preserving visual fidelity.

The dynamic performance further underscores TeleWorld’s advantage. Its WorldScore-Dynamic of 66.73 decomposes into strong Motion Accuracy (53.94), moderate Motion Magnitude (31.55), and high Motion Smoothness (34.18). This profile indicates that motion is plausible, well-regulated, and free of temporal discontinuities—avoiding the under-motion or instability common in baseline systems. This stability stems from TeleWorld’s use of a learned internal state to guide temporal evolution, rather than approximating change locally.

A cross-paradigm analysis shows that TeleWorld bridges a key capability gap: it matches the structural consistency of 3D systems while retaining the conditioning flexibility of video models, and rivals the visual quality of video models while avoiding their typical failures in semantic drift and world collapse. This positions TeleWorld in a previously difficult regime—structurally grounded, flexibly conditioned, and temporally stable generation—supporting its role as a practical step toward interactive, memory-enabled world models.

In summary, the empirical evidence indicates that TeleWorld provides balanced, stable, and scalable world generation capabilities. It does not rely on extreme metric optimization or single-axis specialization. Instead, it demonstrates that a unified model can jointly optimize controllability, consistency, perceptual fidelity, and dynamic behavior. The gains observed in dynamic scores, combined with structural and semantic stability, suggest that TeleWorld is particularly suitable for long-horizon and multi-condition generative tasks. These results identify TeleWorld as a strong candidate for future research directions involving long-range video synthesis, controllable simulation, interactive environments, and world modeling tasks that require coherent spatial-temporal evolution rather than isolated visual quality.


\section{Conclusion}
In summary, TeleWorld is a 18B-parameter model capable of generating high-resolution video (960$\times$1760) in real time at 8 FPS, ranking first on the WorldScores benchmark. It introduces a novel generation‑ reconstruction‑guidance closed‑loop that provides a new solution framework for 4D spatiotemporal world modeling. The model is able to produce long, spatiotemporally consistent 4D scene videos while maintaining persistent 4D memory, offering a valuable reference for subsequent research in world models.

To further speed up scene video generation, we present a scalable and efficient training system that makes Distribution Matching Distillation practical for large-scale auto-regressive video generation models. By decoupling the generator, teacher, and critic across dedicated GPU groups, sharding the generator KV cache via context parallelism, and introducing a carefully balanced pipeline execution schedule, our system overcomes the prohibitive memory and efficiency barriers of applying DMD at the 10B scale and beyond. These system-level optimizations enable DMD training of TeleWorld-18B on a limited GPU budget while sustaining high hardware utilization. In summary, our approach bridges the gap between state-of-the-art distillation techniques and large-scale video diffusion models, unlocking real-time long-horizon video synthesis under practical computational constraints.

\section*{Contributors}
\paragraph{\textbf{Project Leaders:}} Haibin Huang, Chi Zhang, Xuelong Li
\paragraph{\textbf{Core Contributors:}} Yabo Chen, Yuanzhi Liang, Jiepeng Wang
\paragraph{\textbf{Contributors (Listed alphabetically):}} Chengcheng Zhou, Guangce Liu, Haoyuan Wang, Jialun Liu, Junfei Cheng, Junqi Liu, Junyu Zhou, Qizhen Weng, Shiwen Zhang, Tian Li, Tingxi Chen, Wei Li, Weichen Li, Xiaoyan Yang, Xin Zhang, Xuan’er Wu, Xunzhi Xiang, Yuyang Huang, Zicheng Jiang, Zixiao Gu, Zuoxin Li




\bibliography{paper}
\bibliographystyle{authordate1}

\end{document}